\documentclass{article}
\usepackage{spconf,amsmath,graphicx}
\usepackage{enumitem}
\setlist{nosep, leftmargin=14pt}
\usepackage{mwe} 
\usepackage[ruled]{algorithm2e}
\SetKwInput{kwInit}{Init}

\usepackage{pgfplots}
\usetikzlibrary{pgfplots.statistics}
\pgfplotsset{compat=1.16}
\usepackage{subcaption}

\title{Deep Transformers for Fast Small Intestine Grounding in Capsule Endoscope Video}
\name{Xinkai Zhao$^1$, Chaowei Fang$^2$, Feng Gao$^3$, De-Jun FAN$^3$, Xutao Lin$^3$, Guanbin Li$^1$ }
\address{$^1$School of Data and
Computer Science, Sun Yat-Sen University, Guangzhou, China\\
$^2$School of Artifical Intelligence, Xidian University, Xi'an, China \\
$^3$The Sixth Affiliated Hospital, Sun Yat-sen University, Guangzhou, China }

\begin{document}
%
\maketitle
\begin{abstract}
Capsule endoscopy is an evolutional technique for examining and diagnosing intractable gastrointestinal diseases.
Because of the huge amount of data, analyzing capsule endoscope videos is very time-consuming and labor-intensive for gastrointestinal medicalists.
The development of intelligent long video analysis algorithms for regional positioning and analysis of capsule endoscopic video is therefore essential to reduce the workload of clinicians and assist in improving the accuracy of disease diagnosis.
In this paper, we propose a deep model to ground shooting range of small intestine from a capsule endoscope video which has duration of tens of hours. This is the first attempt to attack the small intestine grounding task using deep neural network method.
We model the task as a 3-way classification problem, in which every video frame is categorized into esophagus/stomach, small intestine or colorectum. To explore long-range temporal dependency, a transformer module is built to fuse features of multiple neighboring frames. Based on the classification model, we devise an efficient search algorithm to efficiently locate the starting and ending shooting boundaries of the small intestine. Without searching the small intestine exhaustively in the full video,  our method is implemented via iteratively separating the video segment along the direction to the target boundary in the middle. We collect 113 videos from a local hospital to validate our method. In the 5-fold cross validation, the average IoU between the small intestine segments located by our method and the ground-truths annotated by broad-certificated gastroenterologists reaches 0.945.

\end{abstract}
\begin{keywords}
Capsule Endoscopy, Small Intestine Grounding, Transformer, Convolutional Neural Network
\end{keywords}

\section{Introduction}

\begin{figure}[t]
		\centering
		\includegraphics[width=\linewidth]{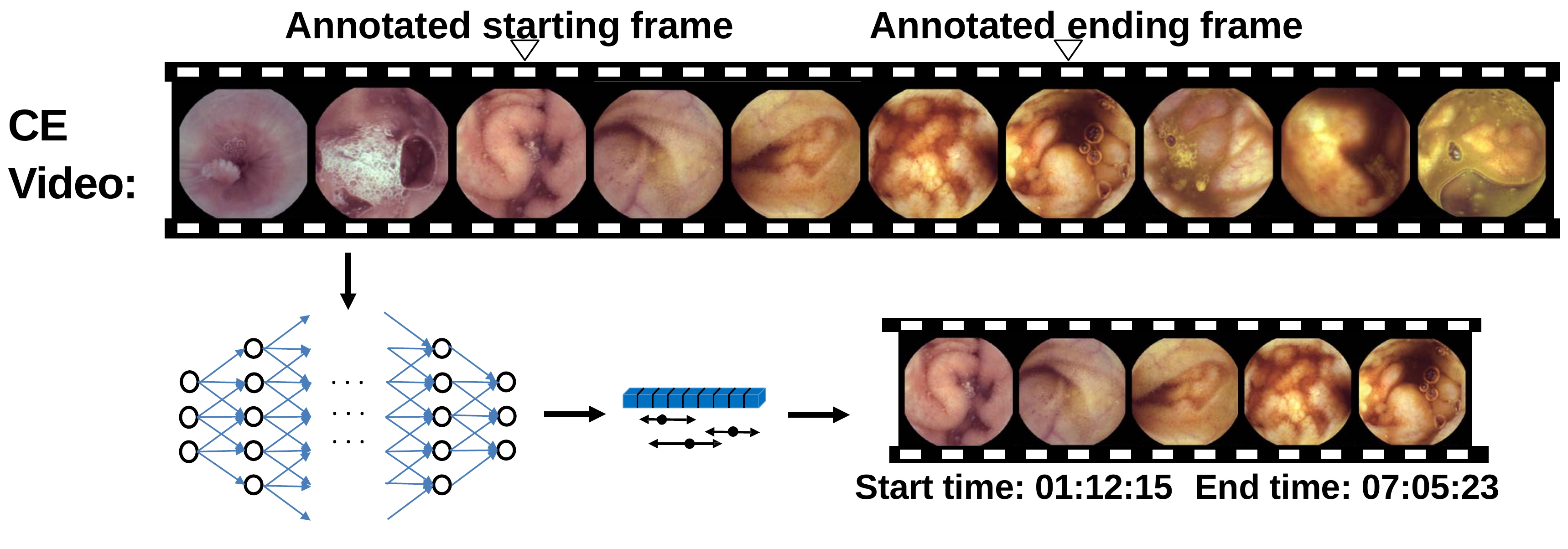}
		\caption{Illustration of the small intestine grounding task. Given a video of capsule endoscope, the grounding task aims at identifying the starting and ending time positions of the small intestine.}
		\label{fig1}
\end{figure}

Capsule endoscope (CE) is a disposable wireless imaging device which is widely used for disease diagnosis in the entire gastrointestinal (GI) tract.
Convenience and noninvasiveness are the main superiorities of capsule endoscope, compared to enteroscopy~\cite{iddan2000wireless,triester2006meta}. In particular, the small intestine mucosa is clearly visualized in videos captured by capsule endoscopes.
There are lots of lesions that may exist in the small intestine, such as Crohn's disease, ulcers, angioectasias, polyps, and bleeding lesions.
With the help of the capsule endoscope, medicalists are able to look into the inner environment of the small intestine and make more accurate disease diagnosis.
However, a full CE video usually occupies over 10 hours and contains over 100,000 frames. It is very time-consuming to screen out lesions from the input video.
Devising automatic machine learning machines to locate the small intestine is valuable as it can greatly reduce the time consumption of subsequent related disease diagnosis.
With the development of artificial intelligence, extensive studies have reported the promising performance of this novel technology for the recognization of various small intestine diseases.
Aoki \emph{et al.}~\cite{aoki2019automatic} use CNNs to detect erosions and ulcerations in CE images.
Ding \emph{et al.}~\cite{ding2019gastroenterologist} use CNNs to identify abnormalities in CE images.
\cite{aoki2020clinical} shows that CNNs can reduce the reading time of endoscopists.
In this paper, we focus on settling the grounding task of the small intestine in a full CE video as illustrated in Fig. \ref{fig1}. This is the first attempt to attack this task via deep learning.

To solve the small intestine grounding problem, we propose an efficient search approach on the basis of a convolutional neural network (CNN).
First of all, we separate the whole video into three components, esophagus/stomach, small intestine and colorectum. A CNN model is built up to fulfill the 3-way video frame classification task. It is composed of three core modules. The backbone of an existing CNN classification such as ResNet~\cite{he2016deep} and DenseNet~\cite{huang2017densely} is regarded as the encoder for extracting feature representations of video frames. The category of every frame is highly related to the categories of preceding and subsequent neighboring frames. For example, the small intestine never comes after the large intestine. Hence, a Transformer module~\cite{vaswani2017attention} is adopted to aggregate features from neighboring frames. Finally, a classification head is used for predicting the category confidence scores. Considering there exists tens of thousands of frames in a full capsule video, it is time consuming to infer the classes of all frames and then seek the small intestine section in brute force. We apply a search algorithm to locate the left and right boundaries of the small intestine. Starting from the middle frame of the whole video, our devised algorithm approaches the target boundary iteratively, reducing the potential boundary positions by half. A dataset containing 113 CE videos is collected from a local hospital, and our method achieves promising grounding results with average IoU of 0.945 under the 5-fold cross validation.


\begin{figure*}[t]
		\centering
		\includegraphics[width=0.9\linewidth]{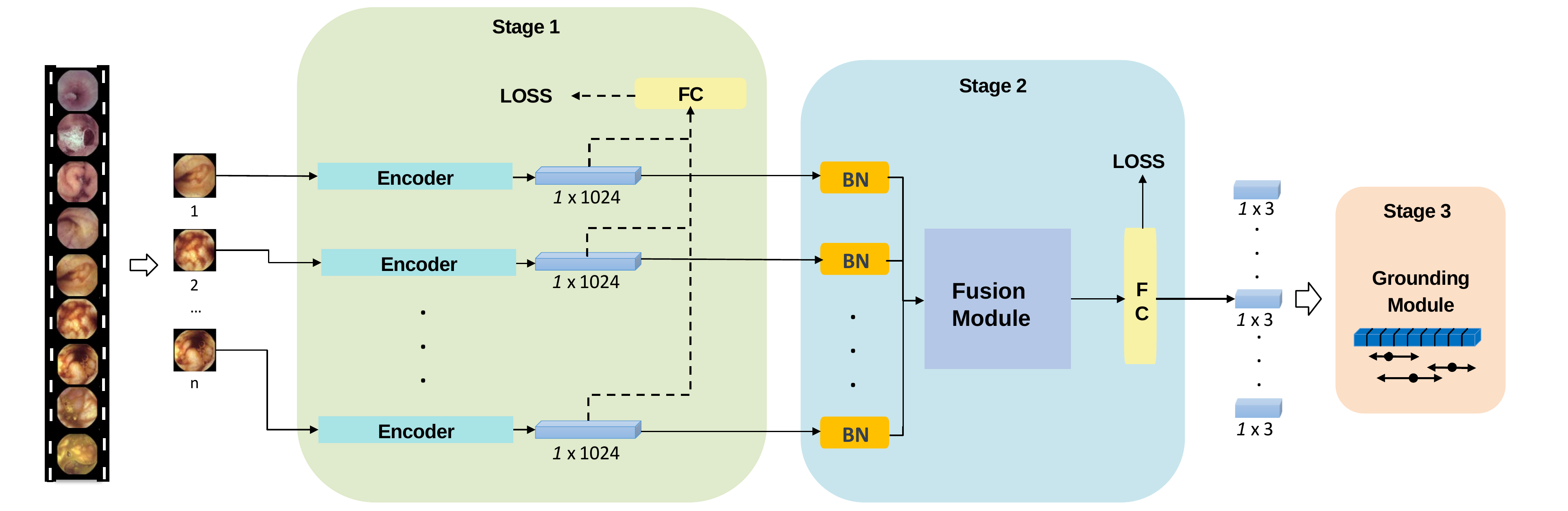}
		\caption{The entire pipeline can be divided into three stages, and after each stage, the trained parameters will be fixed and directly applied to the next stage.}
		\label{fig2}
\end{figure*}

\section{METHOD}
The target of this paper is to ground the small intestine in a full capsule endoscopy video. Provided a video consisting of $T$ frames $\{\mathbf I_t|t=1,\cdots,T\}$, we propose a deep learning based algorithm to locate the starting and ending temporal positions  (denoted as $t_s$ and $t_e$ respectively) of the small intestine. To solve this problem, we first define a 3-way classification problem. Practically, all video frames are classified into 3 categories, including esophagus/stomach, small intestine and colorectum. For frame $\mathbf I_t$, we denote the ground-truth class as $y_t\in\{1,2,3\}$ and the predicted category confidences as $\mathbf p_t\in\mathcal R^3$.
$y_t=1$, $y_t=2$ and $y_t=3$ indicates frame $\mathbf I_t$ belongs to esophagus/stomach, small intestine and colorectum respectively.
To explore inter-frame relationship, deep Transformer is employed to enhance the feature representation of every frame with its neighboring frames. Furthermore, a search algorithm is devised to locate the small intestine. The overall pipeline of our method is shown in Fig.\ref{fig2}.

\subsection{Network Architecture} \label{sec:network}
First of all, we adopt the backbone of an existing classification network as an encoder for extracting the feature representation of every single frame. 
For example, the ResNet~\cite{he2016deep} or DenseNet~\cite{huang2017densely} can be modified as the encoder after removing the last fully connected layer for class probability prediction.
Specifically, we first train a classification network to obtain the features of the frame. Given a video frame $\mathbf I_t$, we denote the extracted feature representation as $\mathbf f_t$.

To capture the temporal inter-frame dependencies, the Transformer module is used to enhance the feature representation of  $\mathbf I_t$ with multiple neighboring frames. Here, the features of $2N+1$ frames $\{\mathbf f_i\}_{i=t-N}^{t+N}$ are
regarded as the input of the Transformer module. Then, we use three fully connected layers to generate the key, query and value vectors for all frames. We denote the key, query, value vector calculated from $\mathbf f_i$ as $\mathbf k_i$, $\mathbf q_i$, and $\mathbf v_i$, respectively. Assume $A$ be the $(2N+1)\times(2N+1)$ attention matrix depicting the relation between every pair of frames. The relation attention between the $i$-th and $j$-th frame is estimated as,
\begin{equation}
A_{i,j} = \mathbf q_i^T \mathbf k_j / \sqrt{m},
\label{formula2}
\end{equation}
where $m$ is the dimension of the feature representation. Every row of $\mathbf A$ is normalized by the softmax function. Then, a new value vector of the $i$-th frame is generated through a matrix multiplication, $\hat{\mathbf v_i}=\sum_{j} A_{i,j} \mathbf v_j$.


To better enhance attention, we use multiple self-attention heads in the transformer. Specifically, 8 heads are adopted to produce 8 new value vectors for every frame, $\{\hat{\mathbf v}_i^m\}_{m=1}^8$. These vectors are concatenated, and compressed into a new feature representation $\hat{\mathbf f_i}$ for the $i$-th frame via a fully connected layer and a residual block consisting of two fully connected layers. The features of $2N+1$ frames, $\{\hat{\mathbf f_i}\}_{i=t-N}^{t+N}$, are fed into another two fully connected layers, yielding the final prediction $\mathbf p_t$ of the $t$-th frame.

The training process is composed of two stages. First, the encoder is optimized as a single frame classification model. A linear layer is used to predict category confidences from the feature of every single frame. In the second stage, the parameters of the encoder is frozen and the parameters of the transformer is optimized. The cross entropy function is adopted to calculate training losses for both stages.

\subsection{Search Algorithm}






To locate the shooting range of the small intestine, we can infer the categories of all frames and then find out the starting and ending frames of the small intestine in brute force. However, this process is inefficient and source-intensive. To increase the efficiency of locating the small intestine with little accuracy loss, we devise a fast and fault-tolerant searching algorithm. With this algorithm, even if some frames are miss-classified, the locating accuracy is still satisfactory.

The procedure for searching the ending position is illustrated in Algorithm \ref{algorithm1}.
Starting from the middle point of the whole video, the searching position $t$ is repeatedly updated to approach the switching point between the left small intestine and the colorectum.
The updating direction depends on the inferred category of the image at current position. If $\mathbf I_t$ belongs to the colorectum, the target position should be later than $t$; otherwise, the target position is earlier than $t$.
The searching interval $d$, which is initialized as $T/2$, is decayed by $\alpha$ after every step.
In order to ensure the algorithm can tolerate some classification errors, the key is the $\alpha$ must be greater than 0.5.

In practice, the confidence value of the inferred class is used to weigh the updating stride. If the inferred class is unconfident, a small stride is adopted to update the searching position.
This is beneficial for increasing the robustness a searching algorithm against incorrect predictions.
We determine the category of $\mathbf I_t$ as, $c_t=\arg\max_{c\in\{1,2,3\}} p_t[c]$ where $p_t[c]$ is the confidence value of the $c$-th class.
The updating stride of the searching position is defines as the following formulation,
\begin{equation}
 \Delta t = \textrm{round}(\alpha*d*(\textrm{max}(p_t[c_t]-\theta,0)+\epsilon)),
\end{equation}
where $\theta$ and $\epsilon$ are constants.

The algorithm of searching the starting position of small intestine can be easily obtained via adjusting the conditions of deciding the updating direction.

\begin{algorithm}[t]
\caption{Algorithm for searching the ending frame of small intestine.}
\SetKwInput{Inputs}{Input}
\Inputs{CE video frames, $\left\{\mathbf I_{i}\right\}_{i=1}^T$.}
\SetKwInput{Output}{Output}
\Output{The ending frame index, $t_e$.}

\kwInit{Frame index, $t = roundint(0.5 * T) $; searching interval, $d = roundint(0.5 * T)$.}

\Repeat{$d < 1$}{
	Use the network in Section \ref{sec:network} to predict the category confidence $\mathbf p_t$, regarding $\{\mathbf I_i\}_{i=t-N}^{t+N}$ as the input;

	$c_t\gets \arg\max_{c\in\{1,2,3\}} p_t[c]$;

	\uIf{$c_t==1$ or $c_t==2$}{
		$t\gets \textrm{roundint}(t+d*(max(p_t[c_t]-\theta, 0)+\epsilon))$
	}
	\Else{
		$t\gets \textrm{roundint}(t-d*(max(p_t[c_t]-\theta, 0)+\epsilon))$
	}
	$d \gets \textrm{roundint}(\alpha * d)$
}
$t_{e} \gets t$
\label{algorithm1}
\end{algorithm}

\section{Experiments}
\subsection{Dataset}
We collect 113 CE videos from the Sixth Affiliated Hospital, Sun Yat-sen University.
These videos are captured by MiroCam\textregistered.
The frame rate is 3fps and the average length is about 11 hours 35 minutes.
The resolution of every video frame is $320 * 320$.
Each video is composed of the shooting ranges of esophagus, stomach, small bowel, and large bowel in chronological order.

We use 5-fold cross validation to evaluate the performance of
the grounding method.
We counted the distribution of labels for the 113 samples. In terms of frame numbers, the esophagus and stomach occupy about 7.2\% of the frames, the small intestine 44.9\%, and the large intestine 47.9\%.

\subsection{Implementation Details}
In the first phase, we choose a learning rate of $10^{-3}$ and a weight decay of $10^{-2}$, and in the second phase, we choose a learning rate of $10^{-5}$ and a weight decay of $10^{-5}$ with $N = 6$ samples to train the fusion module. Adamw optimizer is used in both phases, and the loss function is cross-entropy loss. In the third phase, $\alpha$
is set to 0.9
, $\epsilon$ is set to 0.01, and the threshold $\theta$ is set to 0.5.

\subsection{Quantitative Analysis}

\begin{table}[t]
\centering
\setlength{\tabcolsep}{1mm}{
 \begin{tabular}{ c  c  c  c }
 Method & $IoU$ & $Accuracy$ & $Accuracy*$ \\
 \hline
 VGG & $0.601\pm0.229$ & $0.724$ & $0.570$ \\
 MobileNet & $0.800\pm0.276$ & $0.805$ & $0.571$  \\
 DenseNet & $0.926\pm0.124$ & $0.896$ & $0.872$  \\
 ResNet & $0.932\pm0.092$ & $0.887$ & $0.885$  \\
 DenseNet + LSTM & $0.932\pm0.089$ & $0.897$ & $0.890$  \\
 ResNet + LSTM & $0.944\pm0.065$ & $0.930$ & $0.893$  \\
 DenseNet + TFE & $0.938\pm0.114$ & $0.917$ & $0.891$  \\
 ResNet + TFE & $0.945\pm0.086$ & $0.908$ & $0.911$  \\
\end{tabular}}
\caption{IoU of small-bowel grounding and accuracy of classification}
\label{results}
\end{table}

We present several experiment results comparing other models, i.e. VGG, MobileNet, without inter-frame feature fusion. We also conduct
inner comparisons based two backbones including
DenseNet and ResNet. We put the features of $2N+1$ frames $\{\mathbf F_i\}_{i=t-N}^{t+N}$ into bi-directional long short-term memory (LSTM) and transformer encoder (TFE) to compare the efficacy of fusion methods for classification. The results of multi-class classification are presented in Table \ref{results}. Accuracy in Table \ref{results} means micro-averages and accuracy* means macro-average.
Micro-average and macro-average accuracies are calculated in image level and category level respectively. If the $i$-th class has $n_i$ samples, of which $c_i$ samples are correctly predicted,
\begin{equation}
	accuracy = \frac{\sum_{i=1}^{3}(c_i)}{\sum_{i=1}^{3}(n_i)},~~
	accuracy* = \frac{1}{3} \sum_{i=1}^{3}(\frac{c_i}{n_i})
\end{equation}

Table \ref{cls}
shows the confusion matrix for ResNet and ResNet-TFE, respectively. The results show that the performance is good and balanced. Moreover, the results of ResNet-TFE are slightly better than those of ResNet.



\begin{table}[t]
	\begin{tabular}{ccc|c|c|c|}
	&	& \multicolumn{1}{c}{} & \multicolumn{3}{c}{Ground Truth} \\
	&	& \multicolumn{1}{c}{} & \multicolumn{1}{c}{$1$}  & \multicolumn{1}{c}{$2$}  & \multicolumn{1}{c}{$3$} \\\cline{4-6}
	&	& $1$ & $89.4$ & $4.6$ & $0.3$ \\ \cline{4-6}
	&	ResNet  & $2$ & $8.8$ & $92.9$ & $11.3$ \\\cline{4-6}
	&	& $3$ & $1.9$ & $2.5$ & $88.4$ \\\cline{4-6}
	Prediction &	& \multicolumn{1}{c}{} & \multicolumn{1}{c}{} & \multicolumn{1}{c}{} & \multicolumn{1}{c}{} \\\cline{4-6}
	&	& $1$ & $92.4$ & $2.5$ & $0.3$ \\ \cline{4-6}
	&	ResNet-TFE  & $2$ & $7.2$ & $93.1$ & $7.3$ \\\cline{4-6}
	&	& $3$ & $0.4$ & $4.4$ & $92.3$ \\\cline{4-6}
	\end{tabular}
	\caption{Confusion Matrix of ResNet and ResNet-TFE. The numbers represent
		percentages of images.}
	\label{cls}
\end{table}

\subsection{Grounding Results}

Our proposed algorithm costs 109 seconds averagely to locate the small intestine in full CE videos.
Table \ref{results} also shows the grounding results for each method. Intersection over Union (IoU) is defined as the intersection of the prediction and ground truth divided by the union between them. The results show that the methods devised by us have better performance.

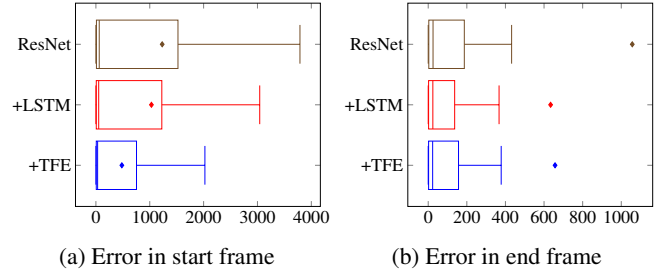
\begin{figure}[t]
\subcaptionbox{Error in start frame}%
      [.49\linewidth]%
      {\begin{tikzpicture}[scale=0.475]
  \begin{axis}
    [
    ytick={3,2,1},
    yticklabels={\Large{ResNet}, \Large{+LSTM}, \Large{+TFE}},
    xticklabels={,\Large{0}, \Large{1000}, \Large{2000}, \Large{3000}, \Large{4000}},
    ]
    \addplot+[
    boxplot prepared={
      median=32,
      upper quartile=754,
      lower quartile=7,
      upper whisker=2024,
      lower whisker=0,
      average = 482
    },
    ] coordinates {};
    \addplot+[
    boxplot prepared={
      median=52,
      upper quartile=1224,
      lower quartile=7,
      upper whisker=3042,
      lower whisker=0,
      average = 1031
    },
    ] coordinates {};
    \addplot+[
    boxplot prepared={
      median=62,
      upper quartile=1524,
      lower quartile=10,
      upper whisker=3789,
      lower whisker=0,
      average = 1231
    },
    ] coordinates {};
  \end{axis}
\end{tikzpicture}}
\hspace*{\fill}
\subcaptionbox{Error in end frame}%
      [.49\linewidth]%
      {\begin{tikzpicture}[scale=0.475]
  \begin{axis}
    [
    ytick={3,2,1},
    yticklabels={\Large{ResNet}, \Large{+LSTM}, \Large{+TFE}},
    xticklabels={,\Large{0}, \Large{200}, \Large{400}, \Large{600}, \Large{800}, \Large{1000}},
    ]
    \addplot+[
    boxplot prepared={
      median=23,
      upper quartile=157,
      lower quartile=2,
      upper whisker=378,
      lower whisker=0,
      average = 657
    },
    ] coordinates {};
    \addplot+[
    boxplot prepared={
      median=24,
      upper quartile=137,
      lower quartile=2,
      upper whisker=367,
      lower whisker=0,
      average = 634
    },
    ] coordinates {};
    \addplot+[
    boxplot prepared={
      median=25,
      upper quartile=187,
      lower quartile=3,
      upper whisker=432,
      lower whisker=0,
      average = 1057
    },
    ] coordinates {};
  \end{axis}
\end{tikzpicture}}
\caption{Error in start and end frame}
\label{boxplot}
\end{figure}

In Fig. \ref{boxplot}, we present the boxplot about the
absolute deviation
between the predicted cutoff point and the ground truth. The results show that the error between the predicted and true values is within 100 frames,
for most of the sample.
The points represents the means, which are between 500 and 1200 because of some outlier points.
And the median is less than 50 for both start and end points.
This indicates that our method achieving promising results in small intestine grounding.

\subsection{Model Interpretability}

\begin{table}[t]
	\begin{tabular}{c c c c}
			 CE & \includegraphics[width=2.15cm]{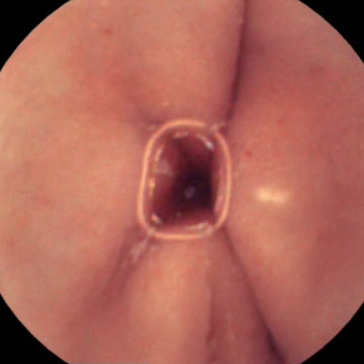} & \includegraphics[width=2.15cm]{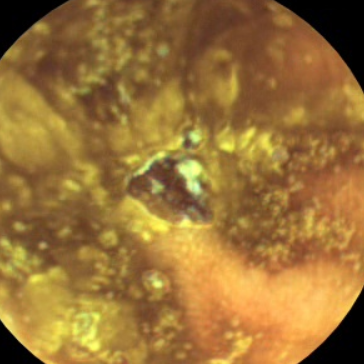} & \includegraphics[width=2.15cm]{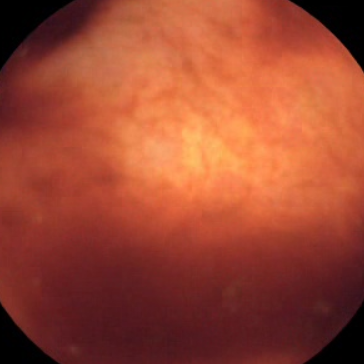} \\
			 HM & \includegraphics[width=2.15cm]{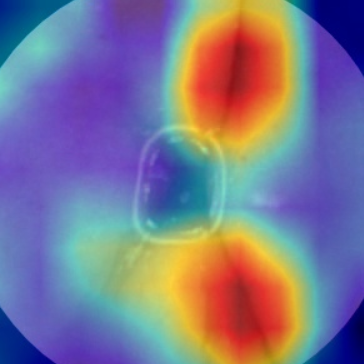} & \includegraphics[width=2.15cm]{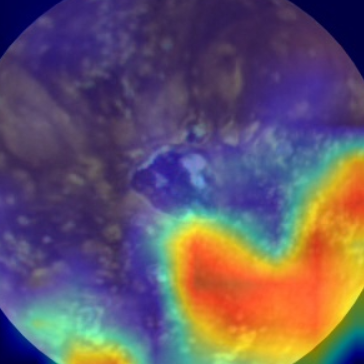} & \includegraphics[width=2.15cm]{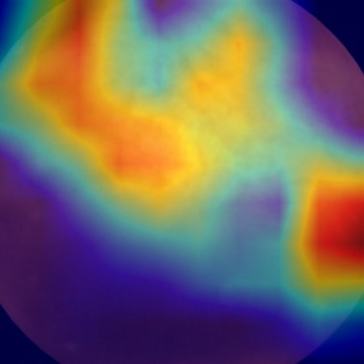} \\
			     & Stomach & Small Bowel & Large Bowel \\
	\end{tabular}
	\caption{Examples of three types of labels and corresponding heat maps (HM) are shown. In these examples, the results for the stomach are more relevant to the folds, the small intestine is judged to be largely dependent on the mucosa, and the large intestine is more concerned with the texture of the inner wall.}
	\label{cam}
\end{table}

In this section, we use class activation map (CAM) to visualize which part of information is discriminative to the classification network.
Table \ref{cam} shows the CAM of some samples, and it can be seen that the network mainly judges the small bowel based on textural information on the mucosa, which is consistent to manual judgment.

\section{DISCUSSION AND CONCLUSION}

We propose a novel algorithm
locate the shooting range of the small intestine in a full CE video.
To the best of our knowledge, it is the first work to solve this problem using deep neural networks. In our approach, we used the transformer encoder for feature fusion between the features of different frames. Besides, we present an efficient algorithm
to search the starting and ending frame of the small intestine based on
a classification network, which achieves good results in terms of efficiency and accuracy.


\bibliographystyle{IEEEbib}
\bibliography{refs}

\section*{Acknowledgments}
This work was supported in part by the Guangdong Basic and Applied Basic Research Foundation (No.2020B1515020048), in part by the National Natural Science Foundation of China (No.61976250, No.61702565, No.62003256). The authors declare that they have no conflict of interest.

\section*{Compliance with Ethical Standards}
This is a numerical simulation study for which no ethical approval was required.

\end{document}